\documentclass[10pt,twocolumn,letterpaper]{article}

\usepackage[pagenumbers]{cvpr} 

\usepackage{graphicx}
\usepackage{caption}
\usepackage{amsmath}
\usepackage{amssymb}
\usepackage{booktabs}
\usepackage{adjustbox} 
\usepackage{multirow}
\usepackage{diagbox}
\usepackage{float}
\usepackage{colortbl}
\usepackage{xcolor}
\usepackage{array}

\usepackage[pagebackref,breaklinks,colorlinks]{hyperref}

\usepackage[capitalize]{cleveref}
\crefname{section}{Sec.}{Secs.}
\Crefname{section}{Section}{Sections}
\Crefname{table}{Table}{Tables}
\crefname{table}{Tab.}{Tabs.}

\def\cvprPaperID{9105} 
\def\confName{CVPR}
\def\confYear{2023}

\begin{document}

\title{Cloud-Device Collaborative Adaptation to Continual Changing Environments \\in the Real-world}


\author{Yulu Gan\textsuperscript{\rm 1}\thanks{Equal contribution}, 
Mingjie Pan\textsuperscript{\rm 1*},
Rongyu Zhang\textsuperscript{\rm 2},
Zijian Ling\textsuperscript{\rm 3},\\
Lingran Zhao\textsuperscript{\rm 1},
Jiaming Liu\textsuperscript{\rm 1},
Shanghang Zhang\textsuperscript{\rm 1}\thanks{Corresponding author}\\
\textsuperscript{\rm 1}Peking University, \textsuperscript{\rm 2}The Chinese University of Hong Kong, Shenzhen, \textsuperscript{\rm 3}Imperial College London\\ 
}

\maketitle

\begin{abstract}

When facing changing environments in the real world, the lightweight model on client devices suffers from severe performance drops under distribution shifts. The main limitations of the existing device model lie in (1) unable to update due to the computation limit of the device, (2) the limited generalization ability of the lightweight model. Meanwhile, recent large models have shown strong generalization capability on the cloud while they can not be deployed on client devices due to poor computation constraints. To enable the device model to deal with changing environments, we propose a new learning paradigm of Cloud-Device Collaborative Continual Adaptation, which encourages collaboration between cloud and device and improves the generalization of the device model.
Based on this paradigm, we further propose an Uncertainty-based Visual Prompt Adapted (U-VPA) teacher-student model to transfer the generalization capability of the large model on the cloud to the device model. Specifically, we first design the Uncertainty Guided Sampling (UGS) to screen out challenging data continuously and transmit the most out-of-distribution samples from the device to the cloud. Then we propose a Visual Prompt Learning Strategy with Uncertainty guided updating (VPLU) to specifically deal with the selected samples with more distribution shifts. We transmit the visual prompts to the device and concatenate them with the incoming data to pull the device testing distribution closer to the cloud training distribution. We conduct extensive experiments on two object detection datasets with continually changing environments. Our proposed U-VPA teacher-student framework outperforms previous state-of-the-art test time adaptation and device-cloud collaboration methods. The code and datasets will be released.

\end{abstract}

\section{Introduction}
\label{sec:intro}

Real-world usually contains various environmental changes along with continual distribution shifts \cite{wang2022continual}. People usually deploy economically lightweight models on devices to boost the scalability and practicability in real-world applications. The lightweight model can suffer severe performance degradation under continual distribution shift \cite{RiccardoVolpi2020ContinualAO, RobertAMarsden2022ContinualUD,wang2022continual, ChristianSimon2022OnGB}. The main challenges are: (1)The poor computational ability of devices. Due to the properties of device infrastructure, the deployed model can not be updated on time, thus lagging its performance for the real world under distribution shift. (2) The limited generalization ability of the lightweight model. Since lightweight models are of relatively small capacity, they can not handle continually changing environments. In contrast, recent large models that are trained on the cloud server show significant generalization ability \cite{RishiBommasani2021OnTO, ChongzhiZhang2021DelvingDI}. While in industry, these large models can not be directly applied due to the limited infrastructure.

Therefore, we enable the device model to tackle real-world environmental changing by proposing a Cloud-Device Collaborative Continual Adaptation paradigm, as shown in Fig .\ref{fig:1} (a). Previous Cloud-Device Collaborative methods only focus on improving model representation on variance in video frames but neglect the model generalization ability for continually changing data distribution. In our new paradigm, we fully exploit the sufficient knowledge of the large cloud model and transfer the continual generalization ability to the device lightweight model.

In particular, we design an Uncertainty-based Visual Prompt Adapted (U-VPA) teacher-student model, which consists of an Uncertainty Guided Sampling (UGS) strategy and a Visual Prompt Learning Strategy with Uncertainty Guided Updating (VPLU). Due to the communication bandwidth constraint, we design the UGS to screen out the most environment-specific samples and decrease the required bandwidth compared with transmitting the whole 
sequence. To leverage the strong generalization ability of large models, we introduce the VPLU to align source-target domain distribution and transfer the representation of the large teacher model to the light student model. The light student model and visual prompts are then delivered to the device, thus alleviating the continuously changing scenarios in the real world.

Experimental results show that our method outperforms the state-of-the-art methods on synthetic and real-world distribution shift datasets, as shown in Fig .\ref{fig:1} (b). 
Besides, we can achieve the same performance as the entire data with fewer reflowed data (42\% of total data). As another benefit, fewer reflowed data reduce the bandwidth pressure of the uplink. As for the downlink, we can deliver the visual prompts (0.43\% of the model's parameters) with almost negligible bandwidth to the device and apply the visual prompt to the input data to improve the performance of the device model by 3.9\% in mAP.

Our contributions can be summarized as follows:

\begin{itemize}
    \item We make the first attempt to deal with continually changing scenarios by proposing a Cloud-Device Collaborative Continual Adaptation paradigm, which aims to transfer the generalization ability from the large cloud model to the lightweight device model. 
    \item We design an Uncertainty-based Visual Prompt Adapted (U-VPA) teacher-student model, which consists of UGS and VPLU. We introduce UGS to screen out the most environment-specific samples and decrease the required bandwidth compared with transmitting the whole sequence. We propose a VPLU to align source-target data distribution and transfer the representation of the large teacher model to the lightweight student model.
    \item Experiments show that our proposed framework and method surpass other state-of-the-art methods and can effectively improve the continuous domain adaptation capability of the device model.
\end{itemize}

\begin{figure*}[t]
\centering
\includegraphics[width=\linewidth]{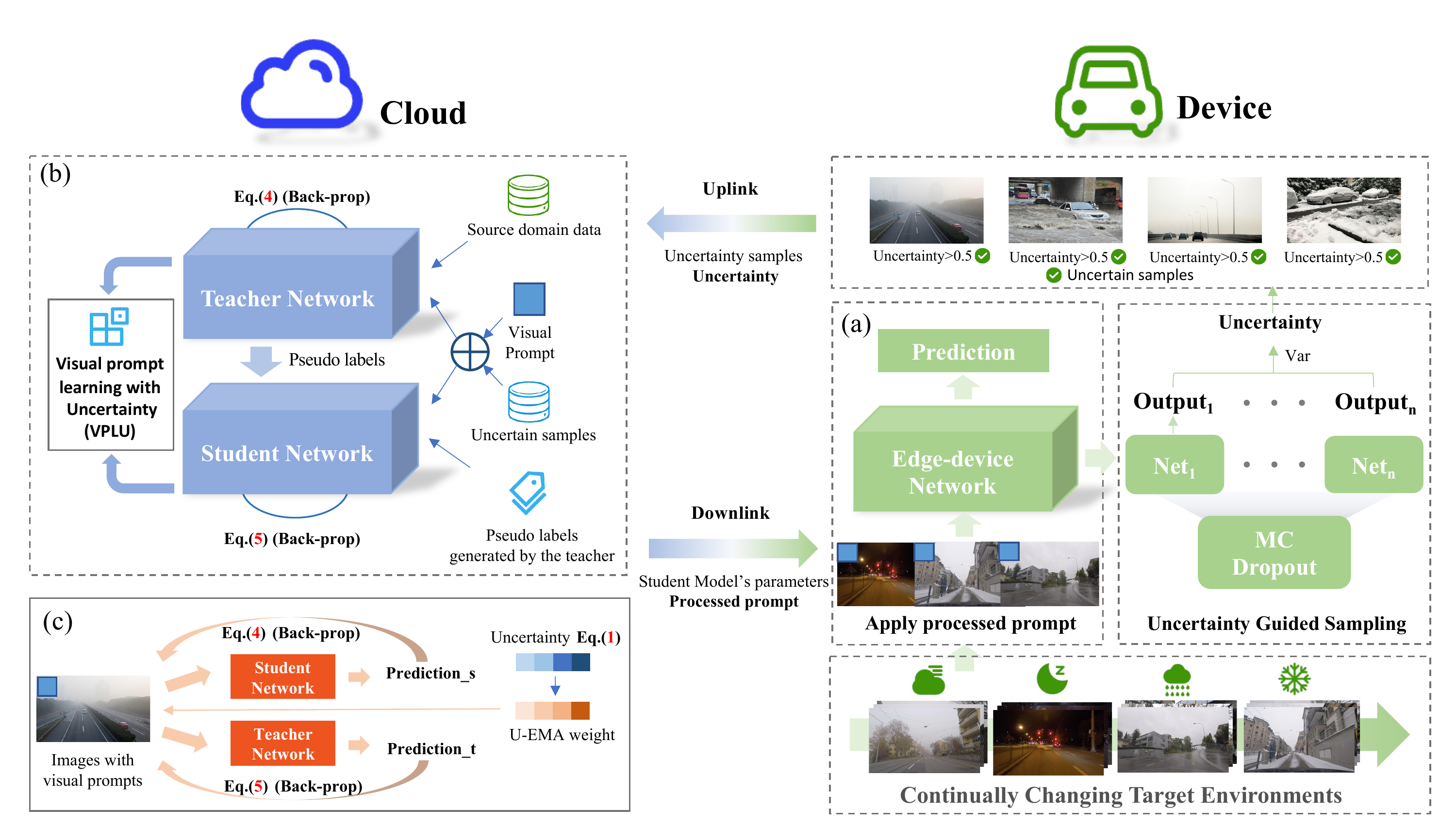}
\vspace{-0.6cm}
\caption{\textbf{The whole framework.} \textbf{(a): Testing on the device}. On the client device, we select data that need to be sent to cloud by Uncertainty Guided Sampling (UGS). These uncertainty samples are then used in the cloud to update the teacher-student models. The visual prompt trained on the cloud will be delivered to the device through the downlink and applied to the input data $x$. Then the reformulated input data will be put into the device model for testing. \textbf{(b) Training on the cloud.} We add our proposed visual prompt to the uncertain samples at the pixel level to get the reformulated images. We then put the reformulated images into the teacher-student model. The teacher-student model and visual prompts are optimized by align loss and supervised loss. \textbf{(c) Visual prompt learning strategy with uncertainty guided sampling (VPLU).} We update the visual prompts by uncertainty-based EMA update. The update weight of the visual prompt will be larger when the uncertainty is higher. The student model is updated using the uncertainty samples and the pseudo labels. }
\label{fig:method}
\vspace{-0.3cm}
\end{figure*}

\section{Related work}
\textbf{Test time adaptation.}
Test-time adaptation(TTA) aims to get better model generalization performance on target data without access to source data. Feature modulation is used in TENT \cite{DequanWang2021TentFT} to optimize BN layers of the input pre-trained model by entropy minimization. SHOT \cite{JianLiang2020DoWR} takes entropy minimization and a diversity regularizer to achieve information maximization. \cite{YuangLiu2021SourceFreeDA , RuiLi2020ModelAU} both take a productive way to enhance model performance on the target domain.

We differ from TTA in that we consider continuously changing domains. Besides, our model is deployed on the device. There will be restrictions on computing resources and uplink and downlink transmission bandwidth. Considering TTA can effectively use the test data to make updates, we leverage TTA to enable the cloud to improve generalization performance continuously.

\textbf{Prompt learning.} Prompt is originally from the natural language process (NLP) and aims to adapt the large-scale pre-trained language model to downstream tasks. Recently, several methods \cite{LimSerNam2022VisualPT, HyojinBahng2022ExploringVP, zhouetal2022coop, zhouetal2022cocoop} are proposed to prove the feasibility of using prompts to adapt the large pre-trained model to downstream tasks in computer vision or vision-language tasks. \cite{LimSerNam2022VisualPT, HyojinBahng2022ExploringVP, ZifengWang2022LearningTP} take prompt as additional input containing a small number of learnable parameters and explore the design of prompts.

Prompt learning aims at mining the pre-trained model's information. We introduce visual prompt tuning to improve the performance of the device model on the current domain, with a negligible computational overhead in our Cloud-Device Collaboration (DCC) framework.



\textbf{Cloud-device collaboration.} Cloud-end collaboration aims to bring benefits to both the cloud and the device jointly. Several previous methods \cite{YipingKang2017NeurosurgeonCI, SandeepChinchali2019NetworkOP,DanielCrankshaw2016ClipperAL,SadjadFouladi2018SalsifyLN} try to unload part or all computation to the cloud to alleviate the computation deficiency on the device. But they do not leverage the collaboration between the cloud and the device.
AMS \cite{MehrdadKhani2020RealTimeVI} and DCCL \cite{JiangchaoYao2021DeviceCloudCL} have considered the collaboration but they do not take the real-world continual distribution shifts into account.
Compared with methods above, our proposed Cloud-Device Collaborative Continual Adaptation paradigm can deal with the real-world environment changes.

In our work, we use Uncertainty Guided Sampling (UGS) and visual prompt for more efficient bandwidth usage. Besides, our model can adapt to different domains during test time. Our teacher model on the cloud and our student model on the device can be improved simultaneously. 

\section{Method}
\label{sec:formatting}
\subsection{Preliminary}
Given a lightweight model $\theta_{device}$ that is deployed on the device, our goal is to perform well on the continually changing target domains $ {D_{t1}, D_{t2} , ... , D_{t T}}$ with limited communication bandwidth, where $D_{t i}={\{(x_i^T)\}}_{i=1}^{N_t}$, and $N_t$ represents the scale of the target domain. The distributions of the target domains are arbitrary since they can change or reoccur over time. We thus introduce a large teacher model $\theta_{tea}$ and lightweight student model $\theta_{stu}$ cloned from $\theta_{device}$, to boost the generalization ability of $\theta_{device}$.
\subsection{Overall framework}

In our Cloud-Device Collaborative Continual Adaptation paradigm, the key idea is to improve the device model's performance continuously by receiving adapted parameters from the cloud server. So we propose Visual Prompt Adapted (U-VPA) teacher-student model. As shown in Fig. \ref{fig:method}, the U-VPA teacher-student model contains Uncertainty Guided Sampling (UGS) and Visual Prompt Learning Strategy with Uncertainty Guided Updating (VPLU).

When perceiving scene changing or distribution shift, poor computational device intends to get help from a cloud server with powerful computing ability. However, due to the communication bandwidth constraint, we must consider the required bandwidth when concurring the distribution shifts. Therefore, we design a Uncertainty Guided Sampling (UGS) to screen out the most environment-specific samples and decrease the required bandwidth compared with transmitting the whole 
sequence. After receiving the learned parameters from the cloud to deal with these samples of relatively large distribution shifts, it can adapt to the new scene when inference.

On the cloud server, due to its strengths in computing, we intend to leverage the strong generalization ability of large models to tackle the received environment-specific and challenging samples. We introduce Visual Prompt Learning Strategy with Uncertainty guided Updating (VPLU) to align source-target domain distribution and transfer the representation of the large teacher model to the light student model. The light student model and visual prompts is then delivered to the device, thus alleviating the continuously changing scenarios in the real-world. 
 

We will discuss the details of UGS and VPLU in Sec. \ref{UGS} and Sec. \ref{VPLU}.

\subsection{U-VPA teacher-student model}

\subsubsection{ Uncertainty Guided Sampling(UGS)}
\label{UGS}
Due to bandwidth limitations, it is not practical for the device to return all data. Therefore, we select images that need to be transmitted based on uncertainty estimation. Inspired by the Dropout methods \cite{YarinGal2015DropoutAA,ChuanGuo2017OnCO}, since we are doing an object detection task, we can obtain n category probabilities $p_i(y|x)$ of the model by dropout method, and we calculate the variance of these n prediction $p_i(y|x)$ as uncertainty.

\begin{equation}
\begin{aligned}
\label{Var_uncertainty}
V_{unc} = (\frac{1}{n}\sum_i^n{(p(y_i|x_i)-\mu)^2})^{\frac{1}{2}}
\end{aligned}
\end{equation}
where $p(y_i|x_i)$ represents the predictions of the input image $x_i$ and $\mu$ is the mean. $V_{unc}$ is the variance of n class probabilities of the model, which represents the uncertainty of the model for the data.

device model filters the input images $x_i$ whose uncertainty is larger than a threshold and sends back to the cloud server.


    

\subsubsection{Visual Prompt Learning Strategy with Uncertainty Guided Updating (VPLU).}
\label{VPLU}

We optimize model on the cloud server by using the data transmitted from the device. It then delivers the learned parameters and processed visual prompt to the device model for inference. We thus introduce VPLU with a large teacher model and light student model, to mitigate variant domain distribution and improve the generalization ability of light model. Note that, teacher model is initialized by the source model pre-trained on source domain. It takes source and target domain as input, while student model only trains on the target domain. 

\textbf{Cross-domain alignment.} In order to mitigate distribution shift, we align the features in the source and target domains. 
Inspired by \cite{YaroslavGanin2015UnsupervisedDA}, we apply feature alignment on teacher model to improve its performance on target domain. We use a domain discriminator to classify data in feature space and try to distinguish which domain the data comes from. Besides, we use a gradient reversal layer to achieve a uniform training process. Source domain data is used to supervise teacher model, avoiding performance degradation.

\textbf{Cross-model transfer.} After improving domain-invariant representation on teacher model, we then transfer its ability in addressing distribution shift and generalization to student model via visual prompt learning and knowledge transferring. 

\textit{Visual Prompt Adaptation(VPA) learning.} Specifically, inspired by the prompt learning in NLP \cite{PengfeiLiu2021PretrainPA}, which can effectively explores the knowledge of source domain, we propose visual prompts as hints to continually transfer teacher model's generalized representation to the student model. 

In particular, visual prompts $\phi$ are learnable parameters. We can obtain the reformulated image $x^*$ by adding visual prompts on the input images $x$ at the pixel level. 

\begin{equation}
\begin{aligned}
x^* = x + \phi,
\end{aligned}
\end{equation}

In our framework, whether in training on server or inference on device, we all use the reformulated image $x^*$ as model's input, instead of origin images $x$.

\textit{Uncertainty-aware Prompt Updation(U-EMA).} 
Visual prompt acts on the input data and participates in the forward process of both teacher and student models, so it can also benefit from the optimization of both teacher and student models in training process. Moreover, to make the prompt update more stable and consider the current model's uncertainty to data, we update the visual prompt using by uncertainty-based exponential moving average(U-EMA). The weight of the exponential moving average is based on the uncertainty estimation as follows:
\begin{equation}
\label{unc_ema_prompt}
\begin{aligned}
\phi_{t} \gets \beta\phi_{t-1} + (1-\beta)\phi_{t},
\end{aligned}
\end{equation}
Note that $\phi_{t}$ represents the parameters of the visual prompt that need to be updated, and $\phi_t$ indicates the visual prompt parameters of the last training step t. We follow \cite{AnttiTarvainen2017MeanTA} to set $\alpha=0.999$. And $\beta = \alpha-V_{unc}$, where $V_{unc}$ is defined in Eq. (\ref{Var_uncertainty}) as the uncertainty of current data. 
Higher uncertainty means a larger distribution shift in the current data. To save more domain-related information, the prompt update weight will be larger at this time. That is, the larger the uncertainty, the greater the weight of the prompt update. We will describe the whole optimization mechanism in the following subsection.

As another essential role of the visual prompt, the processed visual prompt will be sent to the device to improve the adaptability of the device model by applying the visual prompt on input images.

\textbf{Knowlegde transfer.} Along with visual prompt adapted tuning, we align teacher student feature and adopt teacher model to generate pseudo label for student model. It is penalized by Knowledge transfer loss $\mathcal{L}_k$, to promote the consistency between teacher student model representation. Specifically, The teacher model will generate predictions for the reformulated images $x^*$ as pseudo labels. Teacher model has stronger generalization ability, so the generated pseudo labels are of high quality. To further improve the quality of the pseudo labels, we screen those predictions with confidence higher than the threshold as the input to the student model.
Overall, we transfer the knowledge from the teacher model to the student mode by putting selected images $x^*$ and corresponding pseudo labels into the student model for training.

\subsection{Training objectives}
The teacher and student models are jointly optimized on the cloud server, using uncertain samples $x_t^i \in D_t$ returned from device. The source model is trained on source pairs of $\{x_s^i, y_s^i\}^{n_s}_{i=1} \in D_s$ where $x_s^i \in X_s$ and $y_s^i \in Y_s$ are images and labels, respectively.

For the training process of the teacher model, given the reformulated taregt images $x_t^*$, as the Eq. (\ref{teacher_loss}) shown, we use adversal alignment loss \cite{EricTzeng2017AdversarialDD} and supervision loss to optimize the teacher model (including visual prompt $\phi$). 

\begin{equation}
\begin{aligned}\label{teacher_loss}
L_{tea}=-\lambda L_d(G_d(F(x_s^*;\theta_{tea});F(x_t^*;\theta_{tea})), y_d) \\
            + L_{Det}(X(x_s^*, \theta_{tea}),y_s), \\
\end{aligned}
\end{equation}

Note that the first part is adversal alignment loss \cite{EricTzeng2017AdversarialDD}, and another one is the loss function of Faster RCNN \cite{ShaoqingRen2015FasterRT} (using labeled data in source domain $D_s$). $X$ represents the model which predict detection outputs. $L_d$ represent domain classifier. $F$ indicates the feature extractor and $G_d$ is the domain classifier. $\lambda$ is the weight of alignment loss.

For the optimization of the student model, we use pseudo labels $y_t^{p}$ generated by teacher model as supervision:

\begin{equation}
\begin{aligned}\label{teacher_loss}
L_{stu}=L_{Det}(X(x_t^*, \theta_{stu}),y_t^{p}),
\end{aligned}
\end{equation}

The visual prompt benefited from both $L_{tea}$ and $L_{stu}$. And we update the visual prompt by uncertainty-aware exponential moving average as Eq. (\ref{unc_ema_prompt}) shown.

\begin{table*}[htb]
\caption{\label{cityscapes} \textbf{Continual generalization capability on Cityscapes-to-Cityscapes-C.}  Object detection results (mAP@0.5 in \%) on the Cityscapes-to-Cityscapes-C online continual test-time adaptation task. Gain(\%) means the improvement of our method compared with Source-only. We evaluate the four test conditions continually for ten times to evaluate the long-term adaptation performance. All results are evaluated on the FasterRCNN architecture with the largest corruption severity level 5. Our approach surpasses the SOTA method and exhibits significant continual generalization and anti-forgetting abilities.}
\vspace{-0.3cm}
\label{main_table1}
\centering
\setlength\tabcolsep{3pt}
\begin{adjustbox}{width=1\linewidth,center=\linewidth}
\begin{tabular}{l|ccccc|ccccc|ccccc|cc}
\hline
Time      & \multicolumn{16}{c}{$t$ \makebox[14cm]{\rightarrowfill} }                                                                                         &      \\ \hline
Round     & \multicolumn{5}{c|}{1}     & \multicolumn{5}{c|}{5}     & \multicolumn{5}{c|}{10}    & \multicolumn{2 }{c}{All}  \\ \hline
Condition & Fog & Motion & Rain & Snow & Brightness & Fog & Motion & Rain & Snow & Brightness & Fog & Motion & Rain & Snow & Brightness & Mean & Gain\\ \hline
Source-only\cite{ShaoqingRen2015FasterRT} &24.7&9.5&22.4&1.3&29.0&24.7&9.5&22.4&1.3&29.0&24.7&9.5&22.4&1.3&29.0&17.4 & /  \\
TENT-continual\cite{DequanWang2021TentFT} &24.8&9.2&22.5&1.4&29.4&20.0&6.4&19.8&0.8&23.7&9.0&3.8&10.7&0.5&11.4&13.0 & -4.4\\
CoTTA\cite{wang2022continual} &25.1&10.1&22.4&1.3&28.9&21.8&9.5&20.7&4.2&28.2&9.3&3.9&10.1&3.3&17.5&14.7 & -2.7 \\ \hline
Pseudo-Label\cite{DongHyunLee2022PseudoLabelT}
&27.3&17.2&24.7&6.9&33.7&30.1&20.1&26.9&8.0&36.5&31.0&20.9&27.7&7.2&38.2&24.1 & +6.7\\
AMS\cite{MehrdadKhani2020RealTimeVI} &\textbf{28.4}&17.7&26.0&7.6&34.7&31.8&20.4&28.2&7.7&38.0&34.2&21.4&28.9&7.0&39.4&25.1&+7.7 \\
\cellcolor{lightgray} \textbf{Ours(proposed)} &\cellcolor{lightgray} 28.0&\cellcolor{lightgray} \textbf{17.8}&\cellcolor{lightgray} \textbf{26.8}&\cellcolor{lightgray} \textbf{10.0}&\cellcolor{lightgray} \textbf{36.5}&\cellcolor{lightgray} \textbf{37.5}&\cellcolor{lightgray} \textbf{24.2}&\cellcolor{lightgray} \textbf{35.5}&\cellcolor{lightgray} \textbf{14.2}&\cellcolor{lightgray} \textbf{44.0}&\cellcolor{lightgray} \textbf{44.0}&\cellcolor{lightgray} \textbf{27.3}&\cellcolor{lightgray} \textbf{40.9}&\cellcolor{lightgray} \textbf{15.1}&\cellcolor{lightgray} \textbf{50.1}&\cellcolor{lightgray} \textbf{31.0}&\cellcolor{lightgray} \cellcolor{lightgray} \textbf{+13.6} \\ \hline
\end{tabular}
\end{adjustbox}
\end{table*}

\begin{table*}[htb]
\caption{\label{cityscapes} \textbf{Continuous generalization capability on Cityscapes-to-ACDC-Detection.}  Object detection results (mAP@0.5 in \%) on the Cityscapes-to-ACDC-Detection online continual test-time adaptation task. Gain(\%) means the improvement of our method compared with Source-only. We evaluate the four test conditions continually for ten times to evaluate the long-term adaptation performance. All results are evaluated on the FasterRCNN architecture. Our results  outperform the SOTA method, thus verifying the effectiveness of our method on the real distribution shift.}
\vspace{-0.3cm}
\label{main_table2}
\centering
\setlength\tabcolsep{3pt}
\begin{adjustbox}{width=1\linewidth,center=\linewidth}
\begin{tabular}{l|cccc|cccc|cccc|cccc|cc}
\hline
Time      & \multicolumn{16}{c}{$t$ \makebox[12cm]{\rightarrowfill} }                                                                                         &      \\ \hline
Round     & \multicolumn{4}{c|}{1}     & \multicolumn{4}{c|}{4}     & \multicolumn{4}{c|}{7}   & \multicolumn{4}{c|}{10} & \multicolumn{2 }{c}{All}  \\ \hline
Condition & Fog & Night & Rain & Snow & Fog & Night & Rain & Snow & Fog & Night & Rain & Snow & Fog & Night & Rain & Snow & Mean & Gain\\ \hline
Source-only\cite{ShaoqingRen2015FasterRT} &20.5&7.4&14.3&14.2&20.5&7.4&14.3&14.2&20.5&7.4&14.3&14.2&20.5&7.4&14.3&14.2&14.1 &/  \\
TENT-continual\cite{DequanWang2021TentFT} &20.5&7.5&14.3&13.9&20.0&7.2&13.4&13.0&18.6&6.6&12.3&12.1&16.6&6.0&11.1&10.9&12.8&-2.3 \\
CoTTA\cite{wang2022continual} &20.7&7.5&14.7&14.3&20.9&7.4&14.7&14.4&20.4&7.3&13.8&14.4&16.3&6.4&11.7&12.2&13.8 &-0.3 \\ \hline
Pseudo-Label\cite{DongHyunLee2022PseudoLabelT}
&20.9&8.7&15.5&17.6&23.4&10.3&16.9&18.9&24.3&10.9&17.8&19.5&25.1&11.5&17.5&20.0&17.6 & +3.5\\
AMS\cite{MehrdadKhani2020RealTimeVI} &\textbf{21.6}&9.0&\textbf{16.0}&\textbf{17.2}&23.2&10.7&17.2&19.1&24.7&11.4&17.2&19.9&26.1&11.7&17.8&20.5&17.8 &+3.7 \\
\cellcolor{lightgray} \textbf{Ours(proposed)} &\cellcolor{lightgray} 21.1&\cellcolor{lightgray} \textbf{9.3}&\cellcolor{lightgray} 15.7&\cellcolor{lightgray} \textbf{17.2}&\cellcolor{lightgray} \textbf{24.7}&\cellcolor{lightgray} \textbf{11.8}&\cellcolor{lightgray} \textbf{17.8}&\cellcolor{lightgray} \textbf{20.2}&\cellcolor{lightgray} \textbf{26.3}&\cellcolor{lightgray} \textbf{12.0}&\cellcolor{lightgray} \textbf{18.5}&\cellcolor{lightgray} \textbf{20.6}&\cellcolor{lightgray} \textbf{26.1}&\cellcolor{lightgray} \textbf{12.3}&\cellcolor{lightgray} \textbf{18.0}&\cellcolor{lightgray} \textbf{20.8}&\cellcolor{lightgray} \textbf{18.5} &\cellcolor{lightgray} \textbf{+4.4} \\ \hline
\end{tabular}
\end{adjustbox}
\vspace{-0.8em}  
\end{table*}
\section{Experiments}
In Sec.\ref{4.1}, we give the details of the dataset and experiment setting. 
we evaluate the performance of CCA on continual changing scenarios in Sec.\ref{4.2}
We give discussion on our motivations and ablation study in Sec.\ref{4.3}, Sec.\ref{4.4}.

\subsection{Experimental setup}\label{4.1}
\subsubsection{Datasets and settings} 
We evaluate our proposed method on two test time adaptation benchmark tasks for object detection: Cityscapes-to-Cityscape-C and Cityscapes-to-ACDC-Detection dataset.

\textbf{Cityscape-C} is originally created to benchmark robustness tasks \cite{hendrycks2019robustness}. We select the five most relevant corruptions to the autonomous driving scenario (including brightness, motion\_blur, rain, fog and snow). Each corruption dataset contains five levels of severity. The corruptions are applied to images from the validation set of the clean Cityscapes dataset \cite{Cordts2016Cityscapes}. There are 2,500 images in five corruption datasets.

For Cityscapes-to-Cityscapes-C, we follow \cite{mmdetection} to train the pre-trained model on the source domain (Cityscapes). Specifically, we use the SGD optimizer with a learning rate of 1e-2 as the official implementation to train the source model.

\textbf{ACDC-Detection} is created from the Adverse Conditions Dataset (ACDC) \cite{SDV21}. The ACDC dataset shares the same classes with Cityscapes and is collected in various adverse visual conditions: Fog, Night, Rain and Snow. We convert the segmented labels into labels for object detection. We follow CoTTA \cite{wang2022continual} to set our task. We use 400 unlabeled images from each adverse condition for the adaptation. To mimic the scenario in real life where a similar environment might be revisited and to evaluate the continuous generalization capability of our methods, we repeat the same sequence group (of the three conditions) 10 times (i.e., in total 30: Fog $\to$ Rain $\to$ Snow $\to$ Fog ...). This also provides an evaluation of the adaptation performance in the long term.

For the Cityscapes-to-ACDC-Detection task, we use the same way to train the pre-trained model on the source domain (Cityscapes) as the Cityscapes-to-Cityscapes-C task.

\subsubsection{Implementation Details} In this paper, all experiments are conducted with PyTorch. For the Cityscapes-to-Cityscapes-C task and Cityscapes-to-ACDC-Detection task, the batch size is set to 8. We follow \cite{mmdetection} to train the pre-trained model on the source domain (Cityscapes). The number of samples used for uncertainty estimation is set to 10 in the experiment.

We use FasterRCNN with the backbone of ResNet-101 as the large teacher model on the cloud. And we use FasterRCNN with the backbone of ResNet-18 as the lightweight student model (same as the device model). The above large and light models are initialized with corresponding pre-trained models on the source domain (Cityscapes dataset). For the UGS module, we set the uncertainty threshold to 0.008, which filters out 42\% samples in the Cityscape-C dataset. For the VPA module, We set the visual prompts with size 3x200x200, which can be added to the left-up space of the image. The student model's parameters and visual prompts (about 0.4\% of the student model's parameters) will be transmitted to the device.

We uniformly use mAP@0.5(\%) as the evaluation metric. The overall process of test-time adaptation strictly follows CoTTA \cite{wang2022continual}. Our test-time adaptation process will test the data first and then train on the data. We do batch-level updates on the cloud model, and the performance on the device model is taken as the average of the results of all the data.

\textbf{Baselines.} Our method is compared with multiple baselines, including those for test time adaptation (TTA), continual test time adaptation (CTTA) and cloud-end collaboration methods as follows:
\textit{Source-only \cite{ShaoqingRen2015FasterRT} .} Test the model performance on the different target domians using the pre-trained source model (training on the source domain).
\textit{Pseudo-label \cite{DongHyunLee2022PseudoLabelT}.} Use pseudo label generated by the teacher model to supervise the student model.
\textit{Tent-continual \cite{DequanWang2021TentFT}.} Tent is an effective method for the test-time adaptation (TTA). Tent fine-tune the batch normalization layer during training to adapt the model to the target domain.
\textit{CoTTA \cite{wang2022continual}.} CoTTA is an attractive approach for the continual test-time adaptation (CTTA). CoTTA mitigate error accumulation and catastrophic forgetting with weighted averaging, data augmented averaging, and random recovery neurons.
\textit{AMS \cite{MehrdadKhani2020RealTimeVI}.} AMS is an cloud engagement approach. To solve the resource limitation of the device model, AMS do distillation in the cloud, and then transmit the model's parameters to the device in the cloud, the teacher network will not update.

\subsection{Results and Analysis}\label{4.2}
\textbf{Synthetic continual distribution shift.}
We evaluate our proposed method on the synthetic continual distribution shift dataset ( Cityscapes $\to$ Cityscapes-C task). As shown in Table \ref{main_table1}, TENT-continual \cite{DequanWang2021TentFT} has a decreasing performance when facing continuously changing domains. Because TENT does not consider continuous distribution shifts, it uses self-entropy to update the bn layer, which is prone to catastrophic forgetting when facing constant distribution shifts. CoTTA \cite{wang2022continual} is a method designed specifically for continuous test time adaptation. CoTTA applies weighted averaging, data-augmented averaging, and random recovery neurons to overcome catastrophic forgetting and error accumulation. However, when doing the object detection task on the continuously changing domains, the model's performance still decreases, indicating that solutions used by CoTTA are insufficient against catastrophic forgetting. We believe this is caused by the poor generalization abilities of the lightweight device model. Therefore, it is reasonable to introduce the cloud for the Cloud-Device collaborative continual adaptation paradigm.  


Both Pseudo-Label and AMS \cite{MehrdadKhani2020RealTimeVI} use the Cloud-Device collaboration framework,  which transfers the teacher's knowledge to the student on the cloud. The generalization of the large model provides continuous guidance for the student model, which avoids the problem of catastrophic forgetting and error accumulation. (Pseudo-Label and AMS are 6.7\% and 7.7\% higher than Source-only). 

However, pseudo label and AMS do not consider updating the teacher model. Our U-VPA framework jointly optimizes the teacher and student model, which makes our method surpass Source-only by 13.6\%, outperforming all the above methods. This validates that our Cloud-Device collaboration paradigm can effectively improve the continuous domain adaptation ability of the device.


\begin{figure}[t]
\centering
\includegraphics[width=\linewidth]{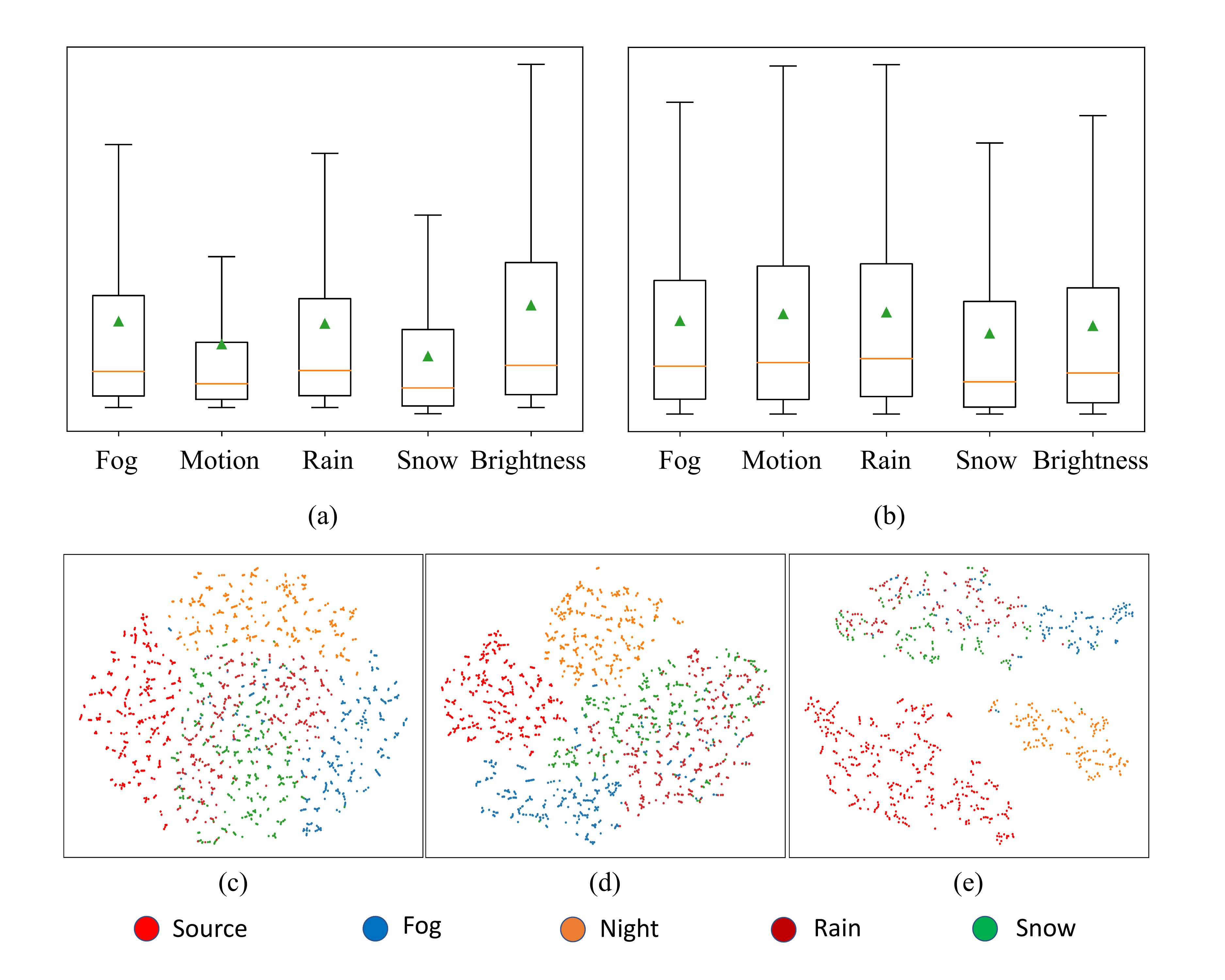}
\vspace{-0.7cm}
\caption{\textbf{(a)-(b): Statistical distribution of confidence and uncertainty.} The results are predicted by the device model on Cityscapes-to-Cityscapes-C tasks. (a): distribution of confidence. (b): distribution of uncertainty. Comparing (a) and (b), the uncertainty value is more stable when facing continuous distribution shift, while the confidence value fluctuates significantly with continuous distribution shift. \textbf{(c)-(e): T-SNE \cite{van2013barnes} visualization of samples in different domains.} (c): all samples. (d): samples filtered by confidence. (e): samples filtered by UGS. Samples filtered by UGS are more environment-specific. They are more worthy of being transmitted to the cloud server for optimizing the teacher-student models.}
\label{t-sne}
\vspace{-0.4cm}
\end{figure}

\begin{table*}[htb]
\centering
\caption{ \textbf{Ablation study.} We conduct ablation studies for proposed Pseudo-Label, VPA, UGS, and U-EMA. All the experiments are done on Cityscapes-to-Cityscapes-C, and the evaluation metric is mAP (\%). The experimental result shows that all these proposed modules positively impact the performance improvement of the devices model when facing continuous domains.}
\label{ablation}
\renewcommand\arraystretch{0.8}
\setlength\tabcolsep{6pt}
\begin{tabular}{lccccccccccc}
\toprule
 &Pseudo-Label&VPA&UGS&U-EMA&Fog&Motion&Rain&Snow&Brightness&Mean&Gain\\\midrule
1 & & & & &24.7&9.5&22.4&1.3&29.0&17.4&/\\
2 &\checkmark& & & &31.0&20.9&27.7&7.2&38.2&24.1&+6.7\\

3 &\checkmark&\checkmark& & &37.1&22.2&32.7&11.2&43.1&27.4&+10.0\\ 
4 &\checkmark&\checkmark&\checkmark& &38.3&24.8&36.5&12.3&46.7&28.3&+10.9\\
5 &\checkmark&\checkmark&\checkmark&\checkmark&44.0&27.3&40.9&15.1&50.1&31.0&+13.6\\\bottomrule

\end{tabular}
\vspace{-0.2cm}
\end{table*}

\textbf{Real-world continual distribution shift.}
We also evaluate our proposed method on the real-world continual distribution shift dataset ( Cityscapes $\to$ ACDC-Detection task). As shown in Table \ref{main_table2}, our method improves by 4.4\% over the source-only method. By comparing the performance change of the same domain in 10 rounds, we can find that our method does not have the problem of catastrophic forgetting. For example, the TENT-continual method faces the problem of catastrophic forgetting. The mAP(\%) drops in the fourth, seventh, and tenth rounds compared to the first round when facing the night domain,  specifically -0.2\%, -0.5\%, -4.1\%. In contrast, our method leads to a continuous performance improvement (e.g., our method's mAP(\%) increases in the fourth, seventh, and tenth rounds, respectively, compared to the first round when facing the Night domain, specifically, +2.5\%, +0.2\%, +0.3\%).
\vspace{-0.1cm}



\begin{table}[!t]
\centering
\renewcommand\arraystretch{1}
\setlength\tabcolsep{2pt}
\vspace{-0.1cm}
\caption{\label{uncertainty} \textbf{Comparing the difference between selecting samples with confidence and uncertainty.} Experiments show that better performance is obtained by choosing the reflowed data model through uncertainty estimation.}
\begin{tabular}{lcccccc}
\toprule
 &Fog&Motion&Rain&Snow&Brightness&Mean\\\midrule
Confidence &39.6&24.9&37.2&14.3&45.1&27.4\\ 
Uncertainty &44.0&27.3&40.9&15.1&50.1&31.0\\\bottomrule
\end{tabular}
\end{table}

\begin{figure}[t]
\centering
\includegraphics[width=0.8\linewidth]{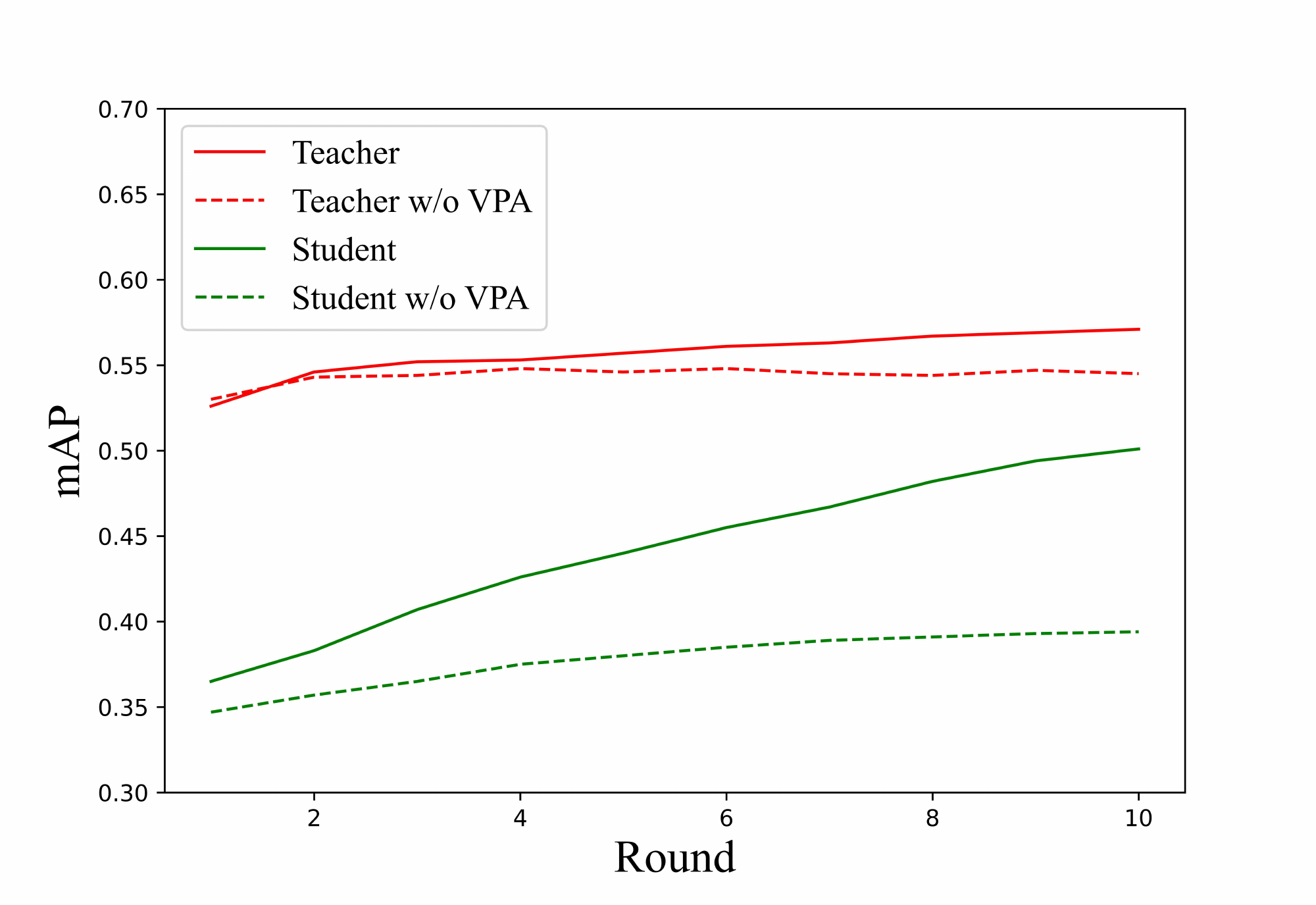}
\vspace{-0.2cm}
\caption{\label{fig:4}\textbf{Detection results (mAP@0.5) of both teacher and student model.} Solid lines are results of models using VPA under the brightness domain. Teacher and student models have jointly improved in continuous domain adaptation. Dotted lines are results without using VPA. The teacher quickly reaches the bottleneck, and student's improvement is slow. The analysis above illustrates that the proposed U-VPA framework enables the teacher-student model jointly improve the continuous domain adaption capability.}
\label{tea_stu}
\vspace{-0.2cm}
\end{figure}

\subsection{Discussion}\label{Analysis and Discussion}\label{4.3}

\textbf{Teacher-student joint optimization with VPA.} Previous approaches (e.g., AMS) focus on improving the student model. However, the student model's performance is correlated with the teacher model. Our method enables the teacher and student models to improve performance together by using the visual prompt as a bridge and using the temporal domain information brought by the visual prompt.

The optimization of the teacher model comes from two modules: (1) feature alignment, (2) visual prompt adaptation(VPA). Our experiments have proved that although both of the above can bring about gains, visual prompt adaptation is the main reason for the success of our framework.

As shown in Fig. \ref{fig:4}, if only feature alignment is used, the performance of the teacher model reaches the bottleneck after the second round, and the student's optimization will be very slow (dotted lines in the figure).

While our method benefits from VPA, the teacher and student models are jointly improved, demonstrating that our proposed U-VPA teacher-student model can optimize the student-teacher model together.

\textbf{Uncertainty Guided Sampling (UGS) strategy.} We demonstrate in Table \ref{unc_few_better_than_total} that using only a small amount of reflowed data can outperform the results using total data by the UGS strategy. Our model's performance using the UGS strategy to select part of the data to transmit into the cloud for training is 31\%. Compared to the model performance of 28.3\% obtained with the total data, there is a 2.7\% improvement. The amount of data using the random picking strategy is the same as that with the UGS strategy. And the UGS strategy has a better performance than random picking. The above shows the effectiveness of the UGS strategy.

We further reveal the advantages of uncertainty when the distribution shifts exist by using boxplot. We find that uncertainty is more reliable compared with the confidence score. According to the results in Fig.\ref{t-sne} (a)-(b), it can be illustrated that when there is a distribution shift, using the UGS strategy to select the reflowed data is better than using select according to confidence. In the presence of distribution shifts, confidence distribution varies significantly, while uncertainty distribution is balanced. Therefore, the samples selected using confidence will face the problem of long-tail distribution, which is not conducive to cloud model training \cite{TongWang2021AdaptiveCS}. Uncertainty can more accurately reflect the difficulty of identifying samples by device without using the training of cloud model and discover the samples that are more worthy of attention.

Besides, we explore the characteristics of the data filtered by different filter strategies. We apply T-SNE to visualize the model features in Fig.\ref{t-sne} (c)-(e). Each point represents a sample filtered by the corresponding strategy. Samples selected by uncertainty have apparent environment specificity, so we can use UGS to filter out more valuable data reflecting domain characteristics and then send it back to the cloud for U-VPA framework training.

\begin{table}[!t]
\centering
\renewcommand\arraystretch{0.8}
\setlength\tabcolsep{12pt}
\vspace{-0.1cm}
\caption{\label{unc_few_better_than_total} \textbf{Effect of uplink transmission of full and partial data on model's performance.} Our method returns partial data (\~42\%) for training but achieves better performance than using full data. This indicates that our UGS strategy reduces the data transmission communication cost and improves the model performance.}
\begin{tabular}{lcccccc}
\toprule
 &25\%&50\%&75\%&100\%\\\midrule
Random &23.5&26.1&27.9 &28.3\\
UGS &28.2&31.0&31.2&/\\
Gain &+4.7&+4.9&+3.3&/\\ 
\bottomrule
\end{tabular}
\vspace{-0.3cm}
\end{table}

\subsection{Ablation study}\label{4.4}

\textbf{Effect of the Uncertainty Guided Sampling (UGS).} Our proposed UGS strategy not only reduces the communication cost of the uplink from the device to the cloud but also improves performance by 0.9\%, as shown in Table \ref{ablation}. Furthermore, as shown in Table \ref{unc_few_better_than_total}, the model using UGS outperforms baseline with only 42\% data for train.

\textbf{Effect of the Visual Prompt Adapt (VPA).} Our proposed VPA teacher-student model uses visual prompts to make the student-teacher model jointly improve performance. As shown in Table \ref{ablation}, compared to the pure pseudo label method, the usage of VPA can increase performance by 3.3\%. Furthermore, we discussed in Sec. \ref{Analysis and Discussion} that VPA could significantly increase the upper limit of the teacher-student framework from challenging continuously changing distribution shift scenarios. The combination of VPA and other modules can further improve performance steadily.

\textbf{Effect of the Uncertainty-aware prompt EMA (U-EMA) update mechanism.} The U-EMA update mechanism we proposed in training the cloud model makes the cloud model pay more attention to the images worthy of attention by dynamically adjusting the prompt weights with uncertainty. As shown in Table \ref{ablation}, U-EMA can bring a 2.7\% performance improvement.





\section{Conclusion}

We propose a Cloud-Device Collaborative Continual Adaptation paradigm to deal with continual changing environments on devices.
We design an Uncertainty Guided Sampling (UGS) strategy to transmit the most out-of-distribution samples from device to cloud. Besides, we design a Visual Prompt Learning Strategy with Uncertanty Guided Updating (VPLU) to transfer the generalization capability of large model on the cloud to the device model.
Experiments of our method on a wide range of continually changing environments show an improvement of 4.4–13.6\% in mAP@0.5 on the object detection task.

{\small
\bibliographystyle{ieee_fullname}
\bibliography{egbib}
}

\end{document}